\pdfoutput=1  

\documentclass[conference]{IEEEtran}
\IEEEoverridecommandlockouts
\usepackage{cite}
\usepackage{amsmath,amssymb,amsfonts}
\usepackage{algorithmic}
\usepackage{graphicx}
\usepackage{textcomp}
\usepackage{xcolor}
\def\BibTeX{{\rm B\kern-.05em{\sc i\kern-.025em b}\kern-.08em
    T\kern-.1667em\lower.7ex\hbox{E}\kern-.125emX}}
\usepackage{array}
\usepackage{relsize}
\usepackage{multirow}
\usepackage{booktabs}
\usepackage{tabularx}
\usepackage{enumitem}
\usepackage{changepage}
\usepackage{adjustbox}
\newcolumntype{P}[1]{>{\centering\arraybackslash}p{#1}}

\usepackage[utf8]{inputenc}
\usepackage[T1]{fontenc}

\usepackage{geometry}
\usepackage{multirow}
\graphicspath{{graphics/}}

\def\paragraph#1{\textbf{#1.} }

\begin{document}
\title{Self-Supervised Pretraining on Satellite Imagery: a Case Study on Label-Efficient Vehicle Detection}

\let\email\relax

\author{\IEEEauthorblockN{1\textsuperscript{st} Jules Bourcier}
\IEEEauthorblockA{\textit{Preligens (ex-Earthcube)} \\
Paris, France\\
\textit{Inria, Univ. Grenoble Alpes,}\\
\textit{CNRS, Grenoble INP, LJK}\\
Grenoble, France\\
\email{jules.bourcier@preligens.com}}
\and
\IEEEauthorblockN{2\textsuperscript{nd} Thomas Floquet}
\IEEEauthorblockA{\textit{Preligens (ex-Earthcube)} \\
Paris, France\\
\textit{MINES Paris - PSL University}\\
Paris, France\\
\email{thomas.floquet@preligens.com}}
\and
\IEEEauthorblockN{3\textsuperscript{rd} Gohar Dashyan}
\IEEEauthorblockA{\textit{Preligens (ex-Earthcube)} \\
Paris, France\\
\email{gohar.dashyan@preligens.com}}
\and
\IEEEauthorblockN{4\textsuperscript{th} Tugdual Ceillier}
\IEEEauthorblockA{\textit{Preligens (ex-Earthcube)} \\
Paris, France\\
\email{tugdual.ceillier@preligens.com}}
\and
\IEEEauthorblockN{5\textsuperscript{th} Karteek Alahari}
\IEEEauthorblockA{\textit{Inria, Univ. Grenoble Alpes,} \\
\textit{ CNRS, Grenoble INP, LJK}\\
Grenoble, France}
\and
\IEEEauthorblockN{6\textsuperscript{th} Jocelyn Chanussot}
\IEEEauthorblockA{\textit{Inria, Univ. Grenoble Alpes,} \\
\textit{ CNRS, Grenoble INP, LJK}\\
Grenoble, France}
}

\maketitle

\begin{abstract}

In defense-related remote sensing applications, such as vehicle detection on satellite imagery, supervised learning requires a huge number of labeled examples to reach operational performances. Such data are challenging to obtain as it requires military experts, and some observables are intrinsically rare. This limited labeling capability, as well as the large number of unlabeled images available due to the growing number of sensors, make object detection on remote sensing imagery highly relevant for self-supervised learning.
We study in-domain self-supervised representation learning for object detection on very high resolution optical satellite imagery, that is yet poorly explored. For the first time to our knowledge, we study the problem of label efficiency on this task.
We use the large land use classification dataset Functional Map of the World to pretrain representations with an extension of the Momentum Contrast framework. We then investigate this model's transferability on a real-world task of fine-grained vehicle detection and classification on Preligens proprietary data, which is designed to be representative of an operational use case of strategic site surveillance.
We show that our in-domain self-supervised learning model is competitive with ImageNet pretraining, and outperforms it in the low-label regime.


\end{abstract}

\section{Introduction}\label{sec:introduction}

Very high resolution (VHR) satellite imagery is one of the key data from which geospatial intelligence can be gathered. It is an essential tool to detect and identify a wide range of objects, on very large areas and on a very frequent basis.
Recently, we have seen the multiplication of available sensors, which has led to a large increase in the volume of data available. This makes it very challenging for human analysts to exploit these data without resorting to automatic solutions.
Deep learning techniques today have been highly effective to perform such tasks.
However, training those models requires very large labeled datasets.
Annotating objects of interest in VHR images can prove to be very costly, being both difficult and time-consuming, and requiring fine domain expertise. In specific contexts such as in geospatial intelligence, the targets can be intrinsically rare, difficult to localize and to identify accurately. This makes the acquisition of thousands of examples impractical, as is typically required for classic supervised deep learning methods to generalize.
Consequently, a major challenge is the development of label-efficient approaches, i.e. models that are able to learn with few annotated examples.

To reduce the number of training samples for difficult vision tasks such as object detection, transfer learning of pretrained neural networks is used extensively. The idea is to reuse a network trained \textit{upstream} on a large, diverse source dataset. ImageNet \cite{russakovsky2015imagenet} has become the de facto standard for pretraining: due to its large-scale and genericity, ImageNet-pretrained models show to be adaptable beyond their source domain, including remote sensing imagery \cite{neumann2020indomain}.
Nonetheless, the \textit{domain gap} between ImageNet and remote sensing domains brings questions about the limitations of this transfer when there are very few samples on the task at hand, e.g. the detection of rare observables from satellite images. To fit the distributions of downstream tasks with maximum efficiency, one would ideally use generic \textit{in-domain} representations, obtained by pretraining on large amounts of remote sensing data. This is infeasible in the remote sensing domain due to the difficulty of curating and labeling these data at the scale of ImageNet. However, imaging satellites provide an ever-growing amount of unlabeled data, which makes it highly relevant for learning visual representations in an unsupervised way.

Self-supervised learning (SSL) has recently emerged as an effective paradigm for learning representations on unlabeled data. It uses unlabeled data as a supervision signal, by solving a \textit{pretext task} on these input data, in order to learn semantic representations. A model trained in a self-supervised fashion can then be transferred using the same methods as a network pretrained on a  \textit{downstream} supervised task. In the last two years, SSL has shown impressive results that closed the gap or even outperformed supervised learning for multiple benchmarks \cite{he2020momentum, chen2020simple, grill2020bootstrap, caron2020unsupervised}.
Recently, SSL has been applied in the remote sensing domain to exploit readily-available unlabeled data, and was shown to reduce or even close the gap with transfer from ImageNet \cite{ayush2021geography, manas2021seasonal, zheng2021self}. Nonetheless, the capacity of these methods to generalize from few labels was not been explored on the important problem of object detection in VHR satellite images.

In this paper, we explore in-domain self-supervised representation learning for the task of object detection on VHR optical satellite imagery. We use the large land use classification dataset Functional Map of the World (fMoW) \cite{christie2018functional} to pretrain representations using the unsupervised framework of MoCo \cite{he2020momentum}. We then investigate the transferability on a difficult real-world task of fine-grained vehicle detection on proprietary data, which is designed to be representative of an operational use case of strategic site surveillance.
Our contributions are:
\begin{itemize}[label=\textbullet]
    \item We apply a method based on MoCo with temporal positives \cite{ayush2021geography} to learn self-supervised representations of remote sensing images, that we improve using (i) additional augmentations for rotational invariance; (ii) a fixed loss function that removes the false temporal negatives in the learning process.
    \item We investigate the benefit of in-domain self-supervised pretraining as a function of the annotation effort, using different budgets of annotated instances for detecting vehicles.
    \item We show that our method is better than or at least competitive with supervised ImageNet pretraining, despite using no upstream labels and 3× less upstream data.
\end{itemize}
Furthermore, our in-domain SSL model is more label-efficient than ImageNet: when using very limited annotations budgets ($\simeq$20~images totalling $\simeq$12k~observables), we outperform ImageNet pretraining by 4~points AP on vehicle detection and 0.5 point mAP on joint detection and classification.

\section{Related work}\label{sec:related_work}

\subsection{Self supervised representation learning}

SSL methods use unlabeled data to learn representations that are transferable to downstream tasks (e.g. image classification or object detection) for which annotated data samples are insufficient. In recent years, these methods have been successfully applied to computer vision with impressive results that closed the gap or even outperformed supervised representation learning on ImageNet, on multiple benchmarks including classification, segmentation, and object detection \cite{he2020momentum, chen2020simple, caron2020unsupervised, grill2020bootstrap}. They commonly rely on a pretraining phase, where a neural network, a \textit{representation encoder}, is trained to solve a pretext task, for which generating labels does not require any effort or human involvement. %
Solving the pretext task is done only for the true purpose of learning good data representations that allow for efficient training on a downstream task of genuine interest.

\subsection{Contrastive learning}

\textit{Contrastive learning} has recently become the most competitive unsupervised representation learning framework, with approaches such as MoCo \cite{he2020momentum}, SimCLR \cite{chen2020simple}, and SwAV \cite{caron2020unsupervised}. Contrastive methods work by attracting embeddings of pairs of samples known to be semantically similar (\textit{positive pairs}) while simultaneously repelling pairs of unlike samples (\textit{negative pairs}). The most common way to define similarity is to use the \textit{instance discrimination} pretext task \cite{dosovitskiy2014discriminative, wu2018unsupervised}, in which positives are generated as random data augmentations on the same image, and negatives are simply generated from different images. Thanks to this pretext task, the encoder learns similar representations for several views of the same object instance in an image and distant representations for different instances. \textit{Momentum Contrast} (MoCo) \cite{he2020momentum} is a strong contrastive method that implements a dynamic dictionary with a queue and a moving-averaged encoder, which enables building a large and consistent dictionary on-the-fly (see section \ref{sec:moco-tp} for more details). In this paper, we adopt a recent \textit{geography-aware} rework of this approach made by \cite{ayush2021geography}.

\subsection{Representation learning in remote sensing}

Building on that success in computer vision, SSL has recently been applied to remote sensing and was shown to reduce or even close the gap with transfer from ImageNet. \cite{jean2019tile2vec} first made use of contrastive learning for remote sensing representation learning and \cite{kang2021} also apply a spatial augmentation criteria on top of MoCo~\cite{he2020momentum}. These works exploit relevant prior knowledge about the remote sensing domain: the assumption that images that are geographically close should be semantically more similar than distant images. Another way of making the learning procedure \textit{geography-aware} is to exploit the image time series that one can get from a given geographic area thanks to the frequent revisit of satellites. This approach is adopted by \cite{ayush2021geography}, that use spatially aligned images over time to construct \textit{temporal positive pairs}. With their \textit{geography-aware} representations learned on the fMoW dataset \cite{christie2018functional}, they improve significantly on classification, segmentation and object detection downstream tasks. However, they do not study label efficiency. In this paper, we apply this method, called MoCoTP, to an operational use-case. We study the label efficiency and bring small extensions to this model, that further improve its performance. In the same vein, \cite{manas2021seasonal} present a pipeline for self-supervised pretraining on uncurated remote sensing data and propose a method that learns representations that are simultaneously variant and invariant to temporal changes. They outperform significantly ImageNet pretraining on classification from few labels on medium resolution images (10m). One can also exploit the multi-spectral and multi-sensor nature of remote-sensing. \cite{stojnic2021self} split multispectral images into two different subsets of channels and use them as augmented (positive) views. \cite{swope2021representation} extend this to multiple sensors, taking subsets of the combination of all bands.
Regarding the domain of pretraining data in remote sensing, \cite{neumann2020indomain} show that the relatedness of the pretraining distribution to the downstream task can improve performances in low-labeled settings. However, they only study supervised pretraining and classification tasks, and show that transfer performance is very dependent on the labeling and data curation quality in the pretraining dataset. Therefore this leaves unresolved the problem of obtaining generic representations with less dependence on labels and this is where SSL can help.

\section{Method}

\begin{figure}
    \centering
    \includegraphics[width=\columnwidth]{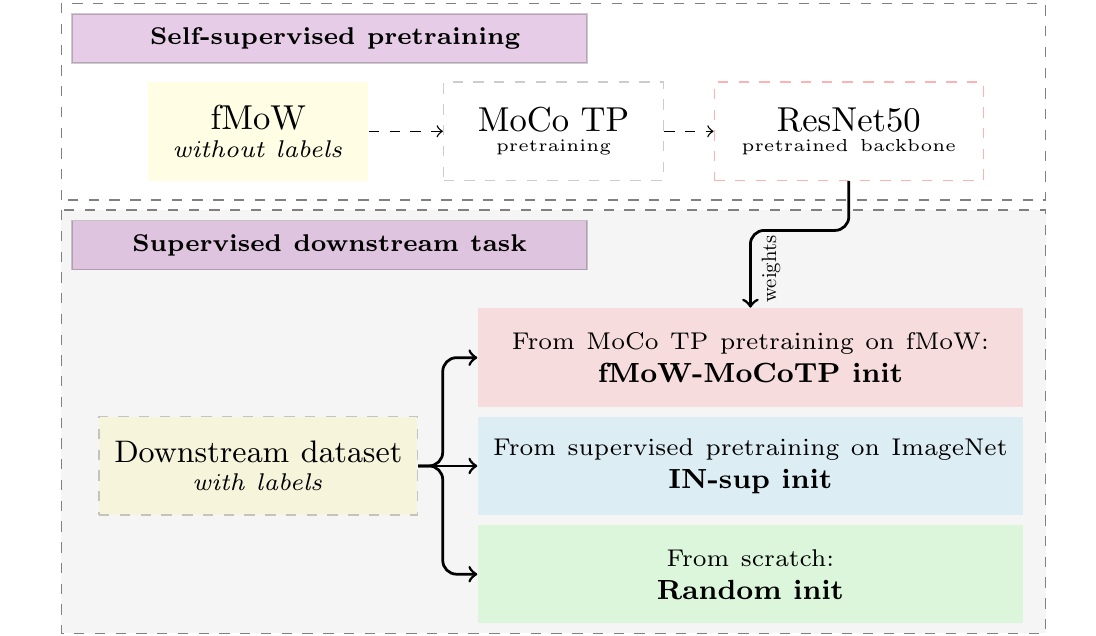}
    \caption{Schematic outline of our method. The top block represents the pretraining phase. The bottom block represents the pretrained weights that are injected into the downstream task model.}
    \label{fig:chart}
\end{figure}

In this section, we detail our approach to explore the applicability of SSL to vehicle detection and classification on optical satellite imagery. The overall procedure is described in Fig.~\ref{fig:chart}. We first pretrain a ResNet-50 backbone on the fMoW dataset \cite{christie2018functional} in an unsupervised way with a recent SSL method, MoCoTP \cite{ayush2021geography}. See section~\ref{sec:moco-tp} for details on the approach and section~\ref{sec:setup-moco-impl} for implementation details.
    
We use the weights of this pretrained backbone on two downstream tasks: (i) fMoW image recognition task: we inject the weights in a classifier, and perform \textit{linear probing} and \textit{finetuning}. See section~\ref{sec:setup-moco} for more details. (ii) Vehicle detection and classification on Preligens proprietary data: we inject the weights in a RetinaNet detector \cite{lin2017focal}, and finetune the entire model. See section \ref{sec:setup-transfer-det-impl} for implementation details.

\subsection{Self-supervised learning with MoCo and Temporal Positives}\label{sec:moco-tp}

We employ the MoCo \cite{he2020momentum} framework for contrastive SSL. The base method we use is from the improved variant MoCo-V2 \cite{chen2020mocov2}.
MoCo learns to match an input \textit{query} $q$ to a \textit{key} $k^+$ (representing the encoded views of the same sample) among a set of negative keys ${k^-}$, using the instance discrimination pretext task \cite{wu2018unsupervised}. It uses a deep encoder (e.g. a ResNet \cite{he2016deep}) to map input image queries and keys to a vector representation space.
Negative keys are extracted with a moving average network (momentum encoder) to maintain consistent representations during training, and are drawn from a memory queue. We refer readers to \cite{he2020momentum} for details on this.
MoCo uses the popular choice of InfoNCE \cite{oord2018representation} for the contrastive loss:
\begin{equation}\label{eq:infonce}
      \mathcal{L}(q, k^+) = -\log\frac
      {e^{(q \cdot k^+/\tau)}}
      {e^{(q \cdot k^+ / \tau)}
       + \sum_{k^-} e^{(q \cdot k^- / \tau)}}
\end{equation}

\noindent
where $\tau$ is a temperature scaling parameter. 

On top of MoCo, we adopt the extension to temporal views proposed in \cite{ayush2021geography}, MoCo with Temporal Positives (MoCoTP). It extends the instance discrimination pretext task to use spatially aligned images from different times as positives. Maximizing similarity between temporal views can provide richer semantic information that extracts persistent scene features over time. The same random augmentations as in MoCo-V2 are also applied on the temporal samples.

\paragraph{Improvements to MoCoTP} Compared to MoCoTP, we further add two modifications to the framework of \cite{ayush2021geography} to improve it:
(i) In addition to the geometric and color perturbations of MoCo-V2, we apply random horizontal flips and rotations by multiples of 90°. Since the data augmentation scheme plays a leading role in contrastive learning \cite{tian2020makes}, we aim to learn representations more suited to overhead images thanks to rotational invariance.
(ii) \cite{ayush2021geography} introduces temporal positives as a drop-in replacement for $q$ and $k^+$ in  \eqref{eq:infonce}. However, this can introduce \textit{false negatives}. Indeed, at each iteration of training, it may happen that the set of negatives ${k^-}$ contains temporal views for samples of the current mini-batch of queries. Such false negatives will cause an incorrect repulsion between the embeddings of similar samples. To what extent this is detrimental to the learned representations depends on the probability of sampling temporal pairs in the training set, as well as on the size of the queue. To avoid the false negatives to interfere with the learning objective, we simply mask out the logits $q \cdot k^-$ in the InfoNCE loss in \eqref{eq:infonce} for every $k^-$ that happens to be a temporal view of $q$.

After being trained on the pretext task for a given number of iterations, the query encoder is extracted and can be transferred to downstream tasks.

\subsection{Transfer to dowstream task}

Following the MoCoTP pretraining on fMoW, we transfer the obtained weights on two downstream tasks: fMoW image recognition task and a real-world use-case, vehicle detection and classification on  Preligens proprietary data. We refer to the downstream training initialized with SSL weights learned on the fMoW dataset as \textbf{fMoW-MoCoTP init}. For each downstream task, we compare fMoW-MoCoTP init with two baselines:
\begin{itemize}
    \item \textbf{IN-sup init}: the backbone has been pretrained on ImageNet in a supervised way;
    \item \textbf{Random init}: the backbone is initialized randomly (i.e. no pretraining).
\end{itemize}

\section{Experimental setup}\label{sec:setup}

\subsection{Pretraining and evaluation on fMoW}\label{sec:setup-moco}

\subsubsection{Dataset}\label{sec:setup-moco-dataset}

For the sake of learning semantic representations in remote sensing, we adopt the fMoW dataset \cite{christie2018functional}, following \cite{ayush2021geography}. fMoW is a public dataset of VHR imagery from Maxar Earth observation satellites, is large-scale with 363,571 training images, and covers 207 countries. It provides images from same locations over time. We apply the self-supervised MoCoTP method for pretraining on fMoW using these available temporal views.
fMoW also includes ground-truths labels for functional land use classification with 62 diverse categories with a long-tailed distribution. We do not use those labels for self-supervised pretraining, but use them downstream to evaluate representations learned directly for the classification of the images seen during pretraining.
Following \cite{christie2018functional} and \cite{ayush2021geography}, we use the fMoW-RGB products for our experiments, which provides 3-bands imagery at 0.5m ground resolution. Preprocessing is applied identically to \cite{christie2018functional} to resize input images to 224×224 pixels.

Ensuing the pretraining stage, the model is evaluated on the land use classification task with the two protocols of linear probing (training a linear classifier on top of frozen features from the pretrained encoder) or finetuning (updating all parameters of the network).
To study the label efficiency of the learned representations, supervised evaluation is performed on 1\%, 10\%, and 100\% of the labeled training data. For 1\% and 10\% labels, we subsample by preserving the base class distribution.  The evaluation metric is the F1-score averaged over classes. Testing is performed on the validation set for comparison with \cite{ayush2021geography}, which consists of 53,041 images.

\subsubsection{Implementation details}\label{sec:setup-moco-impl}

In all our experiments, we use MoCoTP with ResNet-50 for the query and key encoders. Self-supervised pretraining is performed with the following hyperparameters: learning rate of 3e-2 with a cosine schedule, batch size of 256, dictionary queue size of 65536, temperature scaling of 0.2, SGD optimizer with a momentum of 0.9, weight decay of 1e-4. Pretraining is performed for 200 epochs.
For linear probing, we use a learning rate of 1, no weight decay, and only random resized cropping for the augmentations. For finetuning, we use a learning rate of 3e-4 for ResNet weights and 1 for the final classification layer, weight decay of 1e-4, and the same augmentations used for pretraining.
We compare self-supervised pretraining against Random init and IN-sup init, under the different label regimes. Models are trained with cross-entropy loss and evaluated on epoch with the highest top-1 accuracy on the validation set.

\subsection{Transfer to vehicle instance detection}\label{sec:setup-transfer-det}

\subsubsection{Dataset}\label{sec:setup-transfer-det-dataset}

We describe here the Preligens proprietary datasets used for the transfer to object detection.
We call "S" our base dataset. It consists of 204 Maxar WorldView-3 satellite images at 0.3m resolution, and approximately 120k vehicles.

To study the label efficiency, we subsample this base dataset S into smaller datasets, "XS" and "XXS", targeting respectively 50\% and 10\% of the observables present in S. We ensure that XXS is included in XS, so that our datasets follow a "Matriochka" structure which simulates the incremental nature of annotation efforts. The sampling strategy is such that the class distribution is satisfactorily preserved. To perform variance experiments on our results, we perform the sampling thrice and get three different variants of the XS and XXS datasets. We proceed similarly for the S training set by selecting other satellite images that match the class distribution of the initial S dataset.

We keep the same validation and testing sets throughout the experiments, and make sure that the geographical sites of the images are distinct between train and test splits. The training and validation raster images are divided into tiles of 512x512 pixels with an overlap of 128 pixels. Positive tiles (i.e. tiles containing at least one instance) are all kept, but only some negative tiles are kept randomly, to focus training efforts on positive tiles while keeping a fair amount of negative tiles.

Ground truth labels are non-oriented bounding boxes of target observables with their class label. Our classification problem is composed of 8 vehicle categories: \textit{civilian}, \textit{military}, \textit{armored}, \textit{launcher}, \textit{ground support equipment} (\textit{GSE}), \textit{electronics}, \textit{heavy equipment} (\textit{HE}), and \textit{lifting equipment} (\textit{LE}). The datasets statistics are reported in Table~\ref{vehicle-data-table}.

 \begin{table*}[tb] 
  \caption{Data statistics for vehicle training, validation and testing sets. For each of the XXS, XS and S training sets, the reported numbers are the mean number of observables between the different sampled sets.}
  \label{vehicle-data-table}
  \centering
  \adjustbox{max width=\textwidth}{
  \begin{tabular}{l|rrrrrrrrrrrr}
     \toprule
     \centering
    Dataset & Images & Pos. tiles & Neg. tiles & Vehicles & Civilian & Military & Armored & GSE & Launcher & Electronics & HE & LE \\
    \midrule
    XXS & 19 &	597	& 113 & 11,833  & 6,668 & 3,215 & 1,516 & 154 & 158 & 64 & 33 & 23  \\
    XS & 108 &	3,152&	599 & 58,189  & 32,937 & 14,719 & 8,466 & 571 & 732 & 385 & 237 & 139  \\
    S & 204	& 6,438 & 1,231 & 115,617 & 66,504 & 29,332	& 16,148 & 820 & 1364 & 698 & 432 & 319 \\
    \midrule
    Val & 63 & 2,526 & 4,178 & 53,204 & 31,339 & 11,919 & 8,464  & 607 & 412 & 270 & 130 & 101    \\
    Test & 88 & -- &  -- &  32,550 & 19,923 & 7,542 & 3,872 & 361 & 334 & 237 & 184 & 97 \\
    \bottomrule
  \end{tabular}
  }
\end{table*}

\subsubsection{Implementation details}\label{sec:setup-transfer-det-impl}

\textbf{The detection model.} The object detection model used is a RetinaNet \cite{lin2017focal} with a ResNet-50 backbone. The RetinaNet employs a ResNet-FPN \cite{lin2017feature} architecture. %
The backbone is initialized with the pretrained weights learned on fMoW with MoCoTP, and the specific layers of the RetinaNet are initialized randomly. We finetune the RetinaNet model end-to-end, and use the focal loss objective for the classification \cite{lin2017focal}.

\paragraph{Hyperparameters} We selected a learning rate of 1e-4, an Adam optimizer and a batch size of 8.
We used traditional rotations and flips, as well as CLAHE as data augmentations.
Once the validation loss is converged, the epoch selected for the evaluation of the model is the one achieving the best F1-score calculated on the validation set, with a fixed detection score threshold of 0.15.

\paragraph{Evaluation} The F1-score is computed from the precision-recall curves obtained by varying the detection threshold from 0.15 to 0.9. In the following, we refer to the results on the task of vehicle detection (regardless of the class) as \textit{level-1} results, and on the task of joint detection and classification as \textit{level-2} results.
The IoU threshold matching the predictions with the ground truths is set to 0.0, which is operationally relevant for observable counting purposes.
In addition to the F1-score, we also compute the level-1 average precision (AP) and level-2 mean average precision (mAP) which are commonly used metrics to evaluate detection models. Level-1 AP measures the area under the precision-recall curve of level-1 detection, and level-2 mAP is the average of the per class APs.

\section{Results}

\subsection{fMoW classification}

Table~\ref{tab:results-fmow-clf} shows the results of linear probing and finetuning on the 62-class land use classification task of fMoW. With 100\% labels, we see that our improved reproduction of MoCoTP increases performance by 4.36 pts compared to \cite{ayush2021geography} in linear probing and 1.62 pts in finetuning. This shows that the use of the rotation augmentations and the correction of false negatives in the loss function is helpful, especially for linear probing, which closes the gap with finetuning completely. As a note, the sole additional augmentations also improve the baselines Random init and IN-sup init of \cite{ayush2021geography} by 1.29 pts and 0.68 pts respectively.
Moreover, MoCoTP shows impressive label-efficiency: in the semi-supervised settings of 1\% and 10\% labels, we see that it gives respectively 96\% and 87\% of the performance of the network trained with 100\% labels, and surpassing IN-sup init by large margins. These results indicate that MoCoTP is very efficient at learning semantic features from the upstream dataset. Therefore, this is encouraging in order to transfer to a downstream operational task where labeled data are scarce.

\begin{table*}[tb]
    \caption{Results on fMoW classification (F1-score in \%). For 1\% and 10\% labels, the values are 'mean (sd)' across 3 replicates with varying training samples. * denote our improved reproductions as detailed in section \ref{sec:moco-tp}}
    \label{tab:results-fmow-clf}
    \centering
    \begin{tabular}{l|cc|cc|cc}
    \toprule
    ~ & \multicolumn{2}{c|}{1\% labels} & \multicolumn{2}{c|}{10\% labels} & \multicolumn{2}{c}{100\% labels} \\
    ~ & Frozen & Finetune & Frozen & Finetune & Frozen & Finetune \\
    \midrule
    Random init \cite{ayush2021geography} & -- & -- & -- & -- & -- & 64.71 \\
    IN-sup init \cite{ayush2021geography} & -- & -- & -- & -- & -- & 64.72 \\
    fMoW-MoCoTP init\cite{ayush2021geography}       & -- & -- & -- & -- & 64.53 & 67.34 \\
    \midrule
    Random init\,* &  -- &  19.29 (1.65) & -- &   51.87 (0.5) & -- &  65.39 \\
    IN-sup init\,* &  32.41 (0.17) &  39.43 (1.53) &  43.86 (0.07) &  57.32 (0.07) &  50.25 &  66.01 \\
    fMoW-MoCoTP init\,* &  {\bf 60.05 (0.11)} &   {\bf 60.0 (0.43)} &  {\bf 66.15 (0.11)} &  {\bf 66.35 (0.75)} &  {\bf 68.89} &  {\bf 68.96} \\
    \bottomrule
    \end{tabular}

\end{table*}

\subsection{Transfer to vehicle detection}\label{sec:results-transfer-det}

\begin{table*}[tb]
    \caption{Results of each method on each dataset for vehicle detection (\%). The values are 'mean (sd)' across 3 replicates with varying training samples.}
    \centering
    \label{results-level1-table}
    \adjustbox{max width=\textwidth}{
    \begin{tabular}{l | c c c | c c c | c c c}
        \toprule
        \centering
        Metric & ~ & F1 & ~ & ~ & AP & ~ & ~ & mAP & ~ \\
        \midrule
        Training set & XXS & XS & S & XXS & XS & S & XXS & XS & S \\
        \midrule
        Random init & 26.0 (1.9) & 55.1 (2.2) & 74.7 (1.4) & 12.3 (1.9) & 46.9 (2.4) & 75.3 (3.4) & 2.2 (0.3) & 8.4 (0.6) & 16.3 (0.5) \\
        IN-sup init & 61.4 (0.5) & 75.1 (0.8) & \textbf{79.5 (0.6)} & 56.1 (0.9) & 75.6 (0.5) & \textbf{80.9 (0.9)} & 9.3 (0.4) & \textbf{14.9 (0.6)} & \textbf{19.7 (1.1)} \\
        fMoW-MoCoTP init & \textbf{65.1 (0.8)} & \textbf{76.1 (0.4)} & \textbf{79.9 (0.3)} & \textbf{60.1 (1.6)} & \textbf{77.3 (0.1)} & \textbf{81.6 (0.5)} & \textbf{9.7 (0.2)} & \textbf{14.4 (0.4)} & \textbf{19.3 (1.0)} \\
        \bottomrule
    \end{tabular}
    }
\end{table*}

\begin{table*}[tb]
    \caption{\label{results-level2-table}AP per class (\%). Results on GSE and LE are omitted because they are close to zero due to the poor number of examples in the datasets. The values are 'mean (sd)' across 3 replicates with varying training samples.}
    \centering
    
    \adjustbox{max width=\textwidth}{
    \begin{tabular}{l | c c c | c c c | c c c}
        \toprule
        \centering
        Dominant classes & ~ & Civilian & ~ & ~ & Military & ~ & ~ & Armored & ~\\
        \midrule
        Training set & XXS & XS & S & XXS & XS & S & XXS & XS & S \\
        \midrule
        Random init & 14.4 (1.8) & 48.1 (1.2) & 75.4 (4.5) & 3.1 (1.5) & 14.8 (0.6) & 29.1 (3.2) & 0.0 (0.0) & 3.9 (3.5) & 14.9 (7.7) \\
        IN-sup init & 54.2 (1.0) & 75.4 (0.8) & 81.9 (0.7) & \textbf{17.2 (3.8)} & \textbf{25.0 (2.1)} & 32.8 (1.9) & \textbf{0.8 (0.4)} & \textbf{5.7 (0.8)} & \textbf{21.4 (11.4)} \\
        fMoW-MoCoTP init & \textbf{59.9 (1.0)} & \textbf{79.5 (0.8)} & \textbf{83.6 (1.1)} & \textbf{17.9 (0.9)} & \textbf{25.0 (1.4)} & \textbf{35.4 (3.0)} & 0.2 (0.1) & \textbf{6.5 (1.6)} & \textbf{22.4 (11.3)} \\
        \bottomrule
        \toprule
        Rare classes & ~ & Launcher & ~ & ~ & Electronics & ~ & ~ & HE & ~ \\
        \midrule
        Training set & XXS & XS & S & XXS & XS & S & XXS & XS & S \\
        \midrule
        Random init & 0.0 (0.0) & 0.4 (0.3) & 6.0 (2.8) & 0.0 (0.0) & 0.0 (0.0) & 0.1 (0.1) & 0.0 (0.0) & 0.0 (0.0) & 0.0 (0.0) \\
        IN-sup init & \textbf{0.3 (0.2)} & \textbf{9.4 (2.4)} & \textbf{18.8 (4.7)} & 0.0 (0.0) & \textbf{1.5 (0.8)} & \textbf{5.0 (0.2)} & 0.0 (0.0) & \textbf{0.3 (0.5)} & \textbf{1.1 (1.0)} \\
        fMoW-MoCoTP init & 0.1 (0.1) & 4.1 (1.1) & 13.0 (5.7) & 0.0 (0.0) & 0.4 (0.4) & 1.6 (0.4) & 0.0 (0.0) & 0.0 (0.0) & 0.0 (0.1) \\
        \bottomrule
    \end{tabular}
    }
\end{table*}

\begin{figure}
    \centering
    \includegraphics[width=\columnwidth]{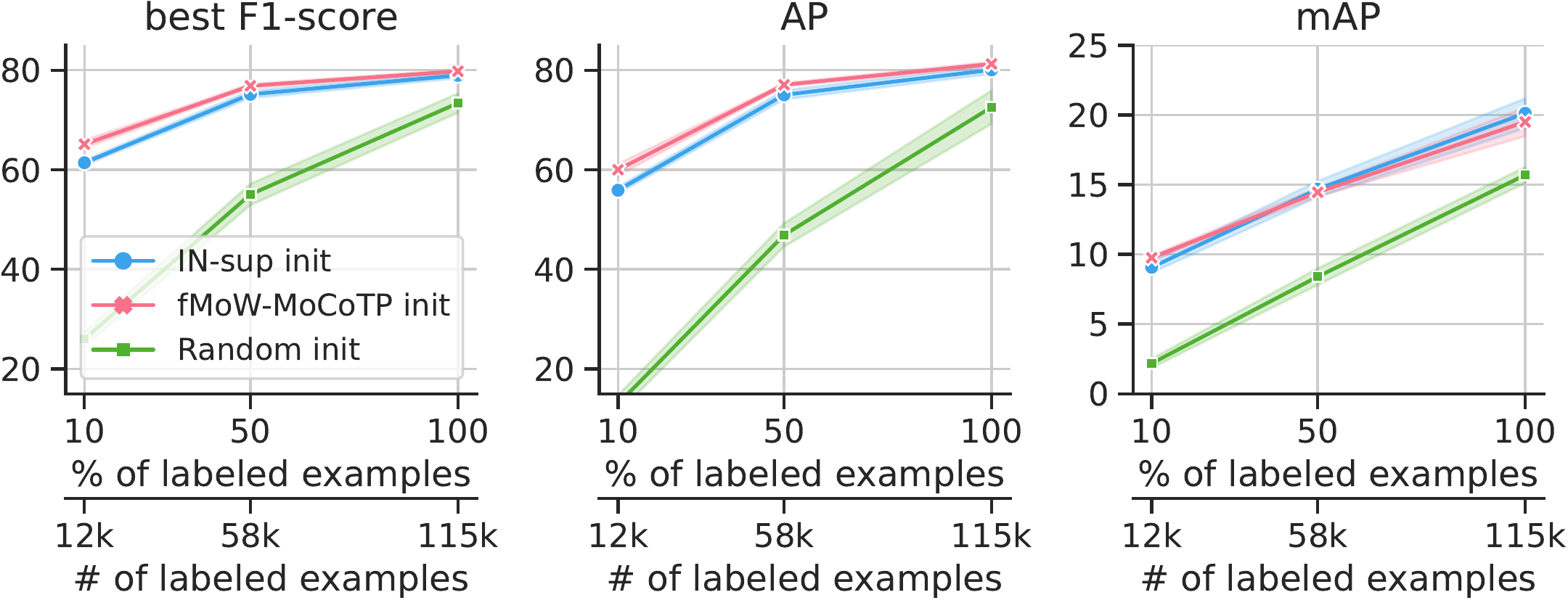}
    \caption{Graphic view of the results of Table \ref{results-level1-table}}
    \label{fig:results_vehicle}
\end{figure}

\paragraph{Label efficiency} Table~\ref{results-level1-table} and Fig.~\ref{fig:results_vehicle} show the results on vehicle detection. The F1-score with fMoW-MoCoTP init is always higher than with IN-sup init or Random init. fMoW-MoCoTP init achieves an F1-score of 65.1\% with only 12k observables on dataset XXS. On dataset XS, with 50\% less examples than on dataset S, fMoW-MoCoTP init is only 3.8~pts below the score obtained on dataset S.
Moreover, the smaller the dataset, the larger the gap between fMoW-MoCoTP init and IN-sup init or Random init: on dataset S, fMoW-MoCoTP init's F1-score is 5.20~pts better than Random init on average, and also 0.40~pts better than IN-sup init, whereas on the XXS dataset, fMoW-MoCoTP init's F1-score is 39~pts better than Random init, and 3.7~pts better than IN-sup init. These results show that self-supervised in-domain pretraining can be competitive with supervised pretraining on ImageNet, and even give better results in low-label regimes.

\paragraph{Dominant vs. rare classes} Table~\ref{results-level2-table} shows the AP in level-2. fMoW-MoCoTP init achieves significantly higher results than Random init on these six classes.
fMoW-MoCoTP init achieves higher AP than IN-sup init on the three dominant classes (Civilian, Military and Armored), that cover $\sim$96.5\% of the vehicles in our datasets. However, IN-sup init outperforms fMoW-MoCoTP init on the Launcher, Electronics and Heavy Equipment classes. These are very rare classes that cover $\sim$2.2\% of the vehicles in our datasets. Since mAP gives equal importance to all classes, this translates into much closer values for mAP scores than level-1 AP scores between fMoW-MoCoTP init and IN-sup init methods, as one can see in Fig.~\ref{fig:results_vehicle}.
This might suggest that fMoW-MoCoTP init is \textit{lazier} and mainly focuses on the dominant classes.
The fMoW dataset used for pretraining contains a long-tailed distribution of semantic categories. One could hypothesize that this leads to representations being more skewed towards over-represented visual concepts than ImageNet, as the latter contains a balanced set of categories, and that such bias may also negatively impact the transfer to under-represented classes downstream. However, further work is needed to provide ground for this hypothesis.

\begin{figure}
\centering
\includegraphics[width=0.95\columnwidth]{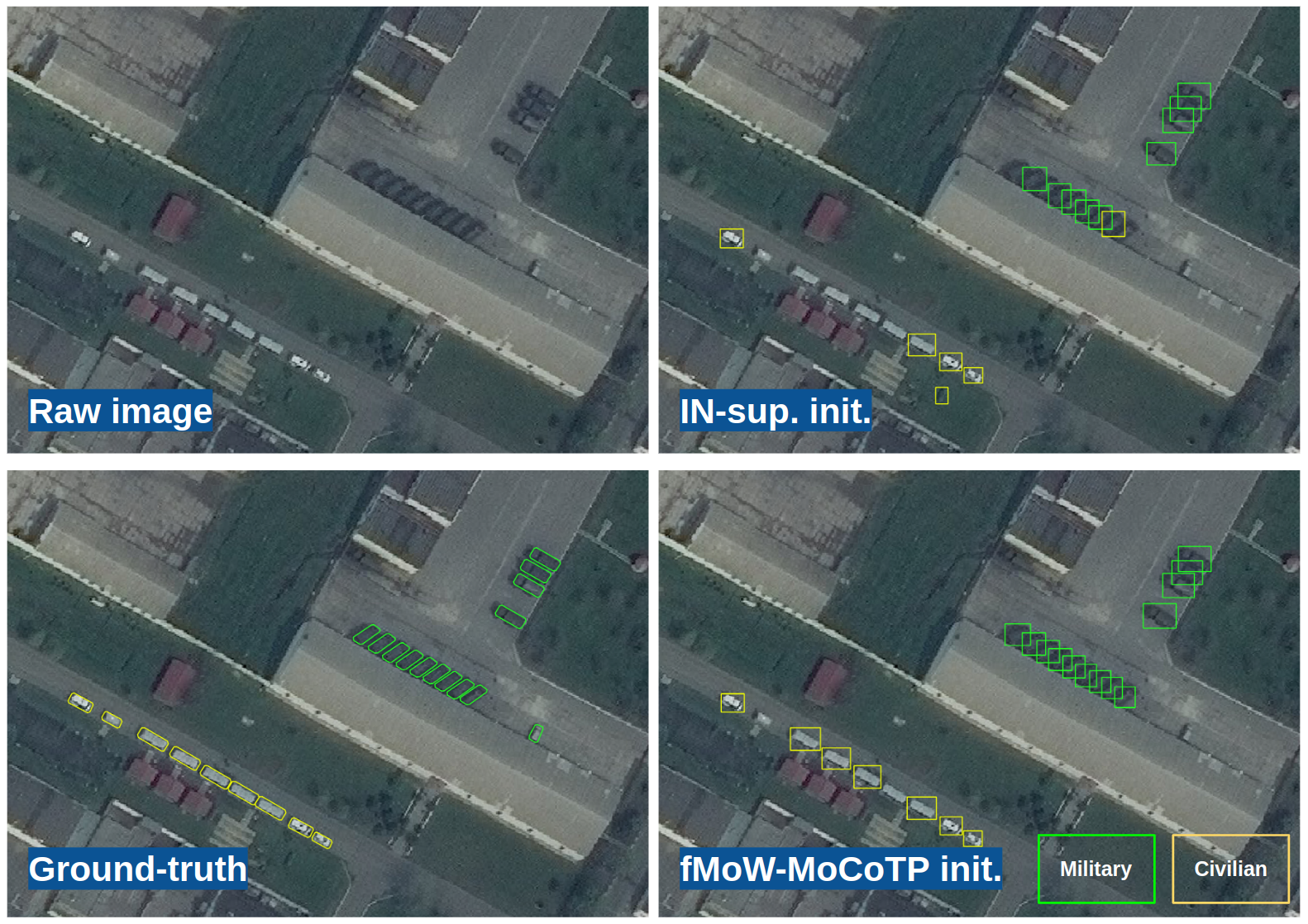}
\caption{Example of cherry-picked predictions. In this particular case, the models have been finetuned on the dataset S.}
\label{}
\end{figure}

\section{Conclusion}\label{sec:conclu}

In this work, we explored the added value of in-domain SSL for a real-world defense-related remote sensing application: vehicle detection and classification on VHR optical satellite imagery. Considering this downstream task, we compared in-domain pretraining on the fMoW dataset with the traditional supervised ImageNet pretraining. We showed that self-supervised pretraining on fMoW is either competitive with or better than supervised ImageNet pretraining, despite using no upstream labels and 3× less upstream data. To study label efficiency, experiments were performed for different downstream dataset sizes, thereby mimicking different annotation budgets. We showed that in-domain SSL pretraining leads to better label efficiency than supervised pretraining on ImageNet. This result is particularly suitable for defense industry use cases, where labeling data is challenging.

Further work could include additional studies such as increasing the amount of in-domain pretraining data, using a vehicle dataset that is more balanced in terms of classes (or deprived of its dominant classes), extending the range of sizes for the downstream dataset, and experimenting other SSL methods, including methods designed specifically for dense downstream tasks like detection such as \cite{wang2021dense, henaff2021efficient, xie2021propagate}.

\subsection*{Acknowledgments}

This work was performed using HPC resources from GENCI–IDRIS (Grant 2021-AD011013097).

\bibliographystyle{splncs04}
\bibliography{references}

\end{document}